\documentclass[aps,prx,groupedaddress,reprint,showpacs,pdftex]{revtex4-1}
\usepackage{graphicx}
\usepackage{tabularx}
\usepackage{dcolumn}
\usepackage{bm}
\usepackage{amsmath}
\usepackage{amssymb}
\usepackage{txfonts}
\usepackage{url}
\usepackage{hyperref}
\usepackage[usenames,dvipsnames]{color}
\definecolor{Gray}{gray}{0.0}
\definecolor{lightGray}{gray}{0.35}
\begin{document}

\title{
  Using reinforcement learning to autonomously identify
  sources of error for agents in group missions
}

\author{
  Keishu Utimula$^1$,
  Ken-taro Hayaschi$^2$, 
  Trevor J. Bihl$^3$,
  Kenta Hongo$^{4}$,
  Ryo Maezono$^{2}$ \\}

\affiliation{\\
  $^1$School of Materials Science, JAIST, 
  Asahidai 1-1, Nomi, Ishikawa 923-1292, Japan\\
  \\
  $^2$School of Information Science, Japan Advanced Institute of Science
  and Technology (JAIST),
  Asahidai 1-1, Nomi, Ishikawa 923-1292, Japan\\
  \\
  $^3$Air Force Research Laboratory,
  Sensors Directorate, WPAFB, OH 45433, USA\\
  \\
  \\
  $^4$Research Center for Advanced Computing
  Infrastructure, JAIST, Asahidai 1-1, Nomi,
  Ishikawa 923-1292, Japan
}

\affiliation{$^{*}$
  mwkumk1702@icloud.com
}

\date{\today}
\begin{abstract}
When agents swarm to execute a mission, 
some of them frequently exhibit sudden failure, 
as observed from the command base. 
It is generally difficult to determine 
whether a failure is caused by actuators 
(hypothesis, $h_a$) or sensors (hypothesis, $h_s$) by solely relying on the communication between the command base and concerning agent. 
However, by instigating collusion between the agents, 
the cause of failure can be identified; in other words, we expect
to detect corresponding displacements for $h_a$ but not for $h_s$. 
{
In this study, we considered the question as to whether 
artificial intelligence can autonomously generate 
an action plan $\bm{g}$ to pinpoint the cause 
as aforedescribed.
Because the expected response to $\bm{g}$ generally 
depends upon the adopted hypothesis
[let the difference be denoted by $D\left(\bm{g}\right)$], 
a formulation that uses $D\left(\bm{g}\right)$ 
to pinpoint the cause can be made. 
Although a $\bm{g}^*$ that maximizes $D\left(\bm{g}\right)$ 
would be a suitable action plan for this task, such an
optimization is difficult to achieve using 
the conventional gradient method, as 
$D\left(\bm{g}\right)$ becomes nonzero 
in rare events such as collisions with other agents, 
and most swarm actions $\bm{g}$ give $D\left(\bm{g}\right)=0$. 
In other words, throughout almost the entire space of $\bm{g}$, 
$D\left(\bm{g}\right)$ has zero gradient, and the gradient method 
is not applicable. 
To overcome this problem, we formulated an action plan 
using Q-table reinforcement learning. 
Surprisingly, the optimal action plan generated via reinforcement 
learning presented a human-like solution to pinpoint the problem 
by colliding other agents with the failed agent. 
Using this simple prototype, we demonstrated 
the potential of applying Q-table reinforcement learning methods 
to plan autonomous actions to pinpoint the causes of failure.
}
\end{abstract}
\maketitle




\section{Introduction}
\label{sec.intro}
The group cooperation of agents 
is an important topic studied 
in the context of autonomous systems.~\cite{2018LEE,2020HU}
Because it is likely for each agent to have 
individual biases in its actuator 
or sensor performance, 
it is an important autonomous ability to 
analyze these inherent biases 
and revise the control plan appropriately 
to continue the group mission. 
Such biases dynamically vary during missions 
with time degradation,
occasionally leading to the failure of some functionality in an agent. 
To ensure appropriate updates to the plan, 
the origins of such biases must be identified. 
\begin{figure*}[htbp]
\includegraphics[width=\hsize]{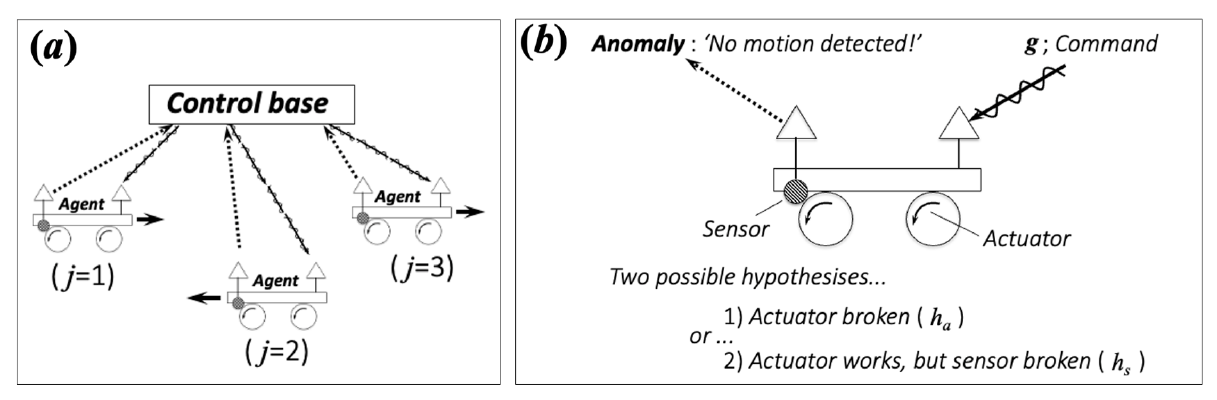}
\caption{
{
Agents perform group actions according to commands communicated from the ``control base`` 
(the figure depicts an example with three agents indexed by $j$). 
The wavy arrow denotes a command signal from the base,
whereas the dotted arrows represent the return signals
from each sensor on each agent [panel (a)]. 
When an anomaly is detected in a return signal, 
two hypotheses - $h_a$ or $h_s$ - can be considered. 
}
}
\label{fig.system}
\end{figure*}

\vspace{2mm}
Suppose that a command base, 
{
which controls a group of agents via 
each command $g_j$~[Fig.~\ref{fig.system}~(a)]
}, 
has detected 
an { anomaly} 
in the position of an agent;
({\it e.g.,} no change in the 
{position} was 
observed). 
There are two possible causes for the observed { anomaly}:
(1) actuator failures (agent is unable to move,) { or} 
(2) sensor failures (agent can move, but 
the move is not captured {by the sensor})
{
~[Fig.~\ref{fig.system}~(b)]
}. 
Depending on the hypothesis, 
[the failure may have occurred in the actuators ($h_a$)
or sensors ($h_s$)], the plan is
{ subsequently} calibrated and updated accordingly.
However, it is generally difficult to identify 
which 
{
hypothesis caused the anomaly 
} 
solely through 
communication between the base and agent.
An intuitive method 
{
to identify the correct hypothesis 
}
is to execute 
{
a collision to the failure agent 
by other agents 
to check whether any displacement 
is observed by the sensor.}
Unless there is a case of sensor failure, such a collision would demonstrate agent displacement. 
Thus, the correct hypothesis can be identified by ``planning a group motion``. 
{
The question then arises as to whether such planning 
can be set up autonomously} as a
``strategy to acquire environmental information``~\cite{2010FRI}.

\vspace{2mm}
{
Such autonomous planning appears to be feasible given 
the following value function.
Suppose that the command $\bm{g}=\left(g_1,g_2,\cdots\right)$ 
is issued from the control base, directing the agent's action 
to specify which of the hypotheses ($h_a$, $h_s$) is supported
~[Fig.~\ref{fig.system}(a)].
This command updates the agent state to 
$\bm{R}\to\bm{\tilde{R}}\left({\bm{g}}\right)$. 
The updated state $\bm{\tilde{R}}$ should be denoted as 
$\bm{\tilde{R}}^{\left(h_l\right)}\left({\bm{g}}\right)$
because it depends on the hypothesis about the state 
before the update ($l=a,s$).
As the expected results differ for different hypotheses, 
the following expression can be used to evaluate the distinction:
$D=\left\|\bm{\tilde{R}}^{\left(h_s\right)}-\bm{\tilde{R}}^{\left(h_a\right)}\right\|$. 
To ensure appropriate planning $\bm{g}$ that involves collisions between agents, 
a non-zero difference $D$ is obtained and the likelihood of 
each hypothesis can be determined.
We must therefore formulate a plan that maximizes
$D=D\left({\bm{g}}\right)$ to ensure a significant difference. 
Accordingly, an autonomous action plan can be formulated to maximize 
$D\left({\bm{g}}\right)$ as a value function. 

\vspace{2mm}
However, this maximization task is difficult to complete
via conventional gradient-based optimization. 
Owing to the wide range of possibilities for $\bm{g}$, 
interactions such as collisions are rare events, 
and for most of the planning phase $\bm{g}$, 
$D\left({\bm{g}}\right)=0$, it is impossible 
to distinguish between hypotheses. 
}
Namely, sub-spaces with finite 
{$D$} 
are sparse
in the overall state space (sparse rewards). 
In such cases, gradient-based optimization 
is insufficient for the task of formulating appropriate action plans
{
because the zero-gradient encompasses the vast majority of the space.} 
For such sparse reward optimization, 
reinforcement learning, which has been thoroughly investigated 
in the applications of autonomous systems ~\cite{2005HUA,2016XIA,2018ZHU,2020HU}, can be used as an 
effective alternative.

\vspace{2mm}
Reinforcement learning
~\cite{2018NAC,2018SUT,2003BAR} 
is becoming an established field 
in the wider context of robotics and system controls.
~\cite{2018PEN,2017FIN} 
Methodological improvements have been studied intensively,
especially by verifications 
on gaming platforms. 
~\cite{2015MNI,2017SIL,2019VIN}
Thus, the topic addressed in this study is 
becoming a subfield known as 
multi-agent reinforcement learning~(MARL).
~\cite{2006BUS,2017GUP,2020STR,2021BIH,2021GRO} 
Specific examples of multi-agent missions include
unmanned aerial vehicles~(UAV)
~\cite{2021BIH,2020STR} and
sensor resource management~(SRM) 
~\cite{2017MAL,1997MAL,2011HER,2021BIH}. 
The objective of this study can also be 
regarded as the problem of
handling non-stationary environments in 
multi-agent reinforcement learning.
~\cite{2020NGU,2017FOE} 
As a consequence of failure, agents are vulnerable
to the gradual loss of homogeneity.
Prior studies have addressed the problem of heterogeneity 
in multiagent reinforcement learning. 
~\cite{2006BUS,2018CAL,2021BIH,2020STR,2021GRO}
The problem of sparse rewards
has also been recognized and discussed 
as one of the current challenges in reinforcement learning.
~\cite{2017WAN,2021BIH}

\begin{figure*}[htbp]
\includegraphics[width=\hsize]{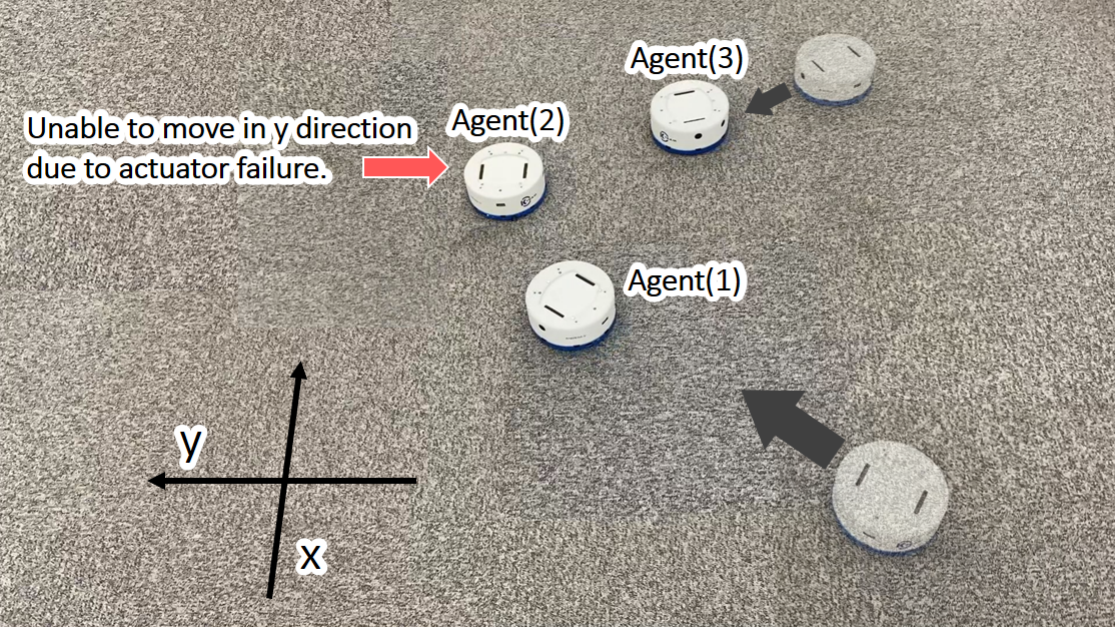}
\caption{
A view of actual machines labelled as Agents \#1-\#3. 
Agent~\#2 is unable to move in the y-direction due 
to actuator failure. 
Agents~\#1 and \#3 are on their way 
to rescue Agent\#2 
(see main text about how the AI decides 
the action plan for the recovery of Agent~\#2). 
}
\label{fig.robots}
\end{figure*}

\vspace{2mm}
As a prototype of such a problem, 
we considered a system composed of 
three agents moving on a ($x,y$)-plane,
administrated by a 
command base to perform a cooperative task (Fig.~\ref{fig.robots}). 
In performing the task, each agent is asked to 
convey an item to a goal post 
individually. 
The second agent (\#2) is assumed to be unable to move along the $y$-direction due to actuator failure.
By quickly verifying tiny displacements in each agent, 
the command base can detect the problem 
occurring in \#2.
However, it cannot attribute the cause to either the actuators or the sensors. 
Consequently, the {control} base sets hypotheses
$h_a$ and $h_s$, and begins planning 
the best cooperative motions {$\bm{g}^*$} to 
classify the correct hypothesis via reinforcement learning. 

\vspace{2mm}
{
Remarkably, the optimal action plan generated 
by reinforcement learning showed a human-like solution 
to pinpoint the problem by colliding other agents 
with the failed agent. 
}
By inducing a collision, the base could identify 
that \#2 is experiencing problems with its actuators rather than 
sensors. 
The base then starts planning group 
motions to complete the conveying task 
considering the limited functionality of \#2. 
We observe that the cooperative tasks are
facilitated by a learning process wherein
other agents appear to 
compensate for the deficiency of \#2 by 
pushing it toward the goal. 
{
In the present study, we employed a simple prototype system 
to demonstrate that reinforcement learning is extremely 
effective in setting up a verification plan 
that pinpoints multiple hypotheses 
for general cases of system failure. 
}

\section{Notations}
\label{sec.notations}
\newcommand{\argmax}{\mathop{\rm arg~max}\limits}
\newcommand{\argmin}{\mathop{\rm arg~min}\limits}
Let the state space for the agents 
be ${\bm R}$. 
For instance, given three agents ($j=1,2,3$) 
situated on a $xy$-plane at positions 
$\left(x_j,y_j\right)$, 
their states can be specified as 
${\bm R}=\left(x_1,y_1,x_2,y_2,x_3,y_3\right)$;
i.e., points in six-dimensional space. 
The state is driven by a command ${\bm g}$ 
according the operation plan generated 
in the command base. 
When ${\bm g}$ is assigned to 
a given ${\bm R}$, 
the state is updated depending on 
which hypothesis $h_l$ is taken, 
each of which restricts ${\bm R}$ 
by individual constraint:
\begin{eqnarray}
{\bm g}: {\bm R} \to \tilde{\bm R}^{\left(h_l\right)}\left({\bm g},{\bm R}\right) \ .
\end{eqnarray}
The difference 
\begin{eqnarray}
D\left({\bm g},{\bm R}\right)
=\sum_{<l,l'>} {\left\|
\tilde{\bm R}^{\left(h_l\right)}\left({\bm g},{\bm R}\right)
-
\tilde{\bm R}^{\left(h_{l'}\right)}\left({\bm g},{\bm R}\right)
\right\|} \ ,
\end{eqnarray}
can then be the measure 
to evaluate performance, and thereby
distinguish between the hypotheses.  
The best operation plan for 
the distinction should therefore be 
determined as 
\begin{eqnarray}
\bm{g}^*
&=&\argmax_{\bm{g}}D\left({\bm g},{\bm R}\right)
\ .
\end{eqnarray}

\vspace{2mm}
{
The naive idea of performing optimization 
using gradient-based methods is insufficient 
owing to the sparseness 
described in the Introduction; 
For $\bm{g}$, 
$D(\bm{g},\bm{R})=0$, the gradient is zero for most of ${\bm R}$ 
because it is incapable of selecting the next update. 
Accordingly, we employed reinforcement learning 
as an alternative optimization approach. 
}

\vspace{2mm}
Reinforcement learning 
assumes the value function 
$\rho\left({\bm R},{\bm g}\right)$,
which measures the gain by taking 
the operation ${\bm g}$ for a state ${\bm R}$.
The leaning process generates a decision that 
maximizes not the temporal 
$\rho\left({\bm R},{\bm g}\right)$,
but also the long-standing benefit
$Q\left({\bm R},{\bm g}\right)$, 
which is the approximate cumulative future gain. 
The benefit $Q$ is evaluated in a 
self-consistent manner (Bellman equation), as~\cite{2018SUT}
\begin{align}
  Q\left({\bm R},{\bm g}\right)
  &=\rho\left({\bm R},{\bm g}\right)\nonumber\\
  &\quad+\sum_{{\bm R'},{\bm g'}}
  F\left(
      \left\{Q\left({\bm R'},{\bm g'}
      \right)\right\},
      \left\{\pi\left({\bm R'},{\bm g'}
      \right)\right\}      
      \right)  
  \ ,
\end{align}
where the second term sums  
all possible states~(${\bm R'}$) and 
actions~(${\bm g'}$) subsequent to 
the present choice $\left({\bm R},{\bm g}\right)$. 
The function 
$F\left(\left\{Q\left({\bm R'},{\bm g'}\right)\right\}\right)$ 
is composed as a linear combination over 
$\left\{Q\left({\bm R'},{\bm g'}\right)\right\}$, 
representing how the contributions get reduced 
over time. 
$\pi\left({\bm R},{\bm g}\right)$
in the second term $F$ describes the policy 
of taking the next decision $\bm g$ at
state $\bm R$.
As explained in Sec.\ref{policy},
$\pi\left({\bm R},{\bm g}\right)$ is
a probability distribution function
with respect to $\bm g$,
reflecting the benefit $Q\left(\bm R,\bm g\right)$ as 
\begin{eqnarray*}
\pi\left(\bm R,\bm g\right)
=P\left(Q\left(\bm R,\bm g\right)\right)
  \ .
\end{eqnarray*}

\vspace{2mm}
$Q\left({\bm R},{\bm g}\right)$ 
is regarded as in Table ($Q$-table) 
with respect to ${\bm R}$ and ${\bm g}$ 
as rows and columns.
At the initial stage,
all values in the table
are set to random numbers
and updated step by step
by self-consistent iterations as follows: 
For the random initial values, 
the temporary decision for initial ${\bm R}_0$ 
is made formally by 
\[
{\bm g}_0\sim \pi\left({\bm R}_0,{\bm g}\right)
\ ,
\]
that is, the sampling by the random distribution
at the initial stage.
Given ${\bm g}_0$, 
a ``point`` $\left({\bm R}_0,{\bm g}_0\right)$ on 
the $Q$-table is updated from the previous 
random value as 
\begin{align}
  Q\left({\bm R}_0,{\bm g}_0\right)
  &=\rho\left({\bm R}_0,{\bm g}_0\right)
\nonumber \\
  &\quad+\sum_{{\bm R'},{\bm g'}}
  F\left(
      \left\{Q\left({\bm R'},{\bm g'}
      \right)\right\},
      \left\{\pi\left({\bm R'},{\bm g'}
      \right)\right\}      
      \right)
  \label{dg}
\end{align}
where $\left\{Q\left({\bm R'},{\bm g'}\right)\right\}$ 
referred from the second term is still 
filled by the random number. 
The operation ${\bm g}_0$ then promotes 
the state to ${\bm R}_0\to {\bm R}_1$. 
Similar procedures are repeated for ${\bm R}_1$:
\begin{eqnarray*}
  {\bm g}_1 &\sim & \pi\left({\bm R}_1,{\bm g}\right)
\\
Q\left({\bm R}_1,{\bm g}_1\right)
&=&\rho\left({\bm R}_1,{\bm g}_1\right)
\nonumber \\
  &&+\sum_{{\bm R'},{\bm g'}}
  F\left(
      \left\{Q\left({\bm R'},{\bm g'}
      \right)\right\},
      \left\{\pi\left({\bm R'},{\bm g'}
      \right)\right\}      
      \right)      
\end{eqnarray*}
As such, the $Q$-table is updated 
in a patchwork manner as sensible values 
replace the initial random numbers. 
Assisted by the neural-network interpolation, 
values are filled for the whole range of the table,
and then converged by the self-consistent 
iteration to get the final $Q$-table. 
In this implementation, 
a user specifies the form of 
$\rho\left({\bm R},{\bm g}\right)$, 
and 
$F\left(\left\{Q\left({\bm R}',{\bm g}'\right)\right\}\right)$, 
providing to the package. 
In this study, we used the 
OpenAI Gym~\cite{2016BRO} package. 

\vspace{2mm}
Denoting the converged table as 
$\bar Q\left({\bm R},{\bm g}\right)$,
we fix the policy as 
\begin{eqnarray}
\bar\pi\left(\bm R,\bm g\right)
=P\left(\bar Q\left(\bm R,\bm g\right)\right)
\ ,
\label{policyProb}
\end{eqnarray}
to generate the series of 
operations for state updates: 
\begin{eqnarray}
  \bar{\bm g}_0
      &\sim & \bar\pi\left({\bm R}_0,{\bm g}\right)
  \nonumber \\
  &&\bar{\bm g}_0 : {\bm R}_0\to {\bm R}_1
  \nonumber \\
  \nonumber \\
  \bar{\bm g}_1
      &\sim & \bar\pi\left({\bm R}_1,{\bm g}\right)
  \nonumber \\
  &&\bar{\bm g}_1 : {\bm R}_1\to {\bm R}_2
            \nonumber \\
           \cdots && \ .
\label{sequence}
\end{eqnarray}

\section{Experiments}
\label{sec.experiments}
The workflow required to achieve the mission for 
the agents, as described in Sec.~Introduction, proceeds as 
follows: 
\begin{itemize}
\item [[0a]]~To determine if there are errors  
found in any of the agents, 
the base issues commands 
to move all agents by tiny displacements 
(and consequently, \#2 is found to have 
an error). 
\item [[0b]]~Corresponding to 
each possible hypothesis ($h_a$ and $h_s$), 
the virtual spaces 
$\left\{{\bm R}^{\left(h_l\right)}\right\}_{l=a,s}$ 
are prepared by applying each constraint. 
\item [[1]]~
Reinforcement learning ($Q_\alpha$) 
is performed at the command base using the virtual space, 
generating ``the operation plan $\alpha$`` 
to distinguish the hypotheses. 
\item [[2]]~
The plan $\alpha$ is performed by the agents. 
The command base compares the observed trajectory 
with that obtained in the virtual spaces in Step [1]. 
In the process, the hypothesis that yields the closest 
trajectory to that observed is identified as 
accurate ($h_a$). 
\item [[3]]~
By taking the virtual space 
${\bm R}^{\left(h_a\right)}$ 
as the identified hypothesis, 
another learning $Q_\beta$ is performed 
to get the optimal plan $\beta$ for the original mission 
(conveying items to goal posts). 
\item [[4]]~
Agents are operated according to the plan $\beta$. 
\end{itemize}
All learning processes and operations  
are simulated on a Linux server. 
The learning phase is the most time-intensive, 
requiring approximately 3h using a single processor 
without any parallelization to complete. 
For the learning phase, we implemented the 
PPO2 (proximal policy optimization, version2) 
algorithm~\cite{2015SCH} from the 
OpenAI Gym~\cite{2016BRO} library. 
Reinforcement learning $(Q_{\alpha})$ 
was benchmarked on the MLP~(multilayer perceptron) and 
LSTM~(long-short time memory) network structures, 
with performance compared between them. 
We did not conduct specific tuning 
for the hyperparameters as a default setting,
{
  as shown in Table ~\ref{table.hp}.
}
However, it has been pointed out that hyperparameter 
optimization (HPO) can significantly 
improve the performance of reinforcement learning. 
~\cite{2018HEN,2020STR,2020BIH,2012SNO,2015DOM,2021BIH,2020YOU}
The comparison indicates that MLP performs better, with possible reasons given in the third paragraph of \S\ref{sec.discussions}. 
The results described herein
Were obtained by the MLP network structure.  
Notably, LSTM also generated almost identical agent
behaviors to those exhibited by the MLP (possible reasons are given in 
the Appendix, \S\ref{res.lstm}.

\begin{table}[hbtp]
  \caption{PPO2 hyperparameters used in training.}
  \label{table.hp}
  \centering
  \begin{tabular}{ll}
    \hline
    Parameter  & Value  \\
    \hline \hline
    gamma           & 0.99     \\
    n\_steps        & 128      \\
    ent\_coef       & 0.01     \\
    learning\_rate  & 0.00025  \\
    vf\_coef        & 0.5      \\
    max\_grad\_norm & 0.5      \\
    lam             & 0.95     \\
    nminibatches    & 4        \\
    noptepochs      & 4        \\
    cliprange       & 0.2      \\
    \hline
  \end{tabular}
\end{table}

\begin{figure*}[htbp]
  \includegraphics[width=\hsize]{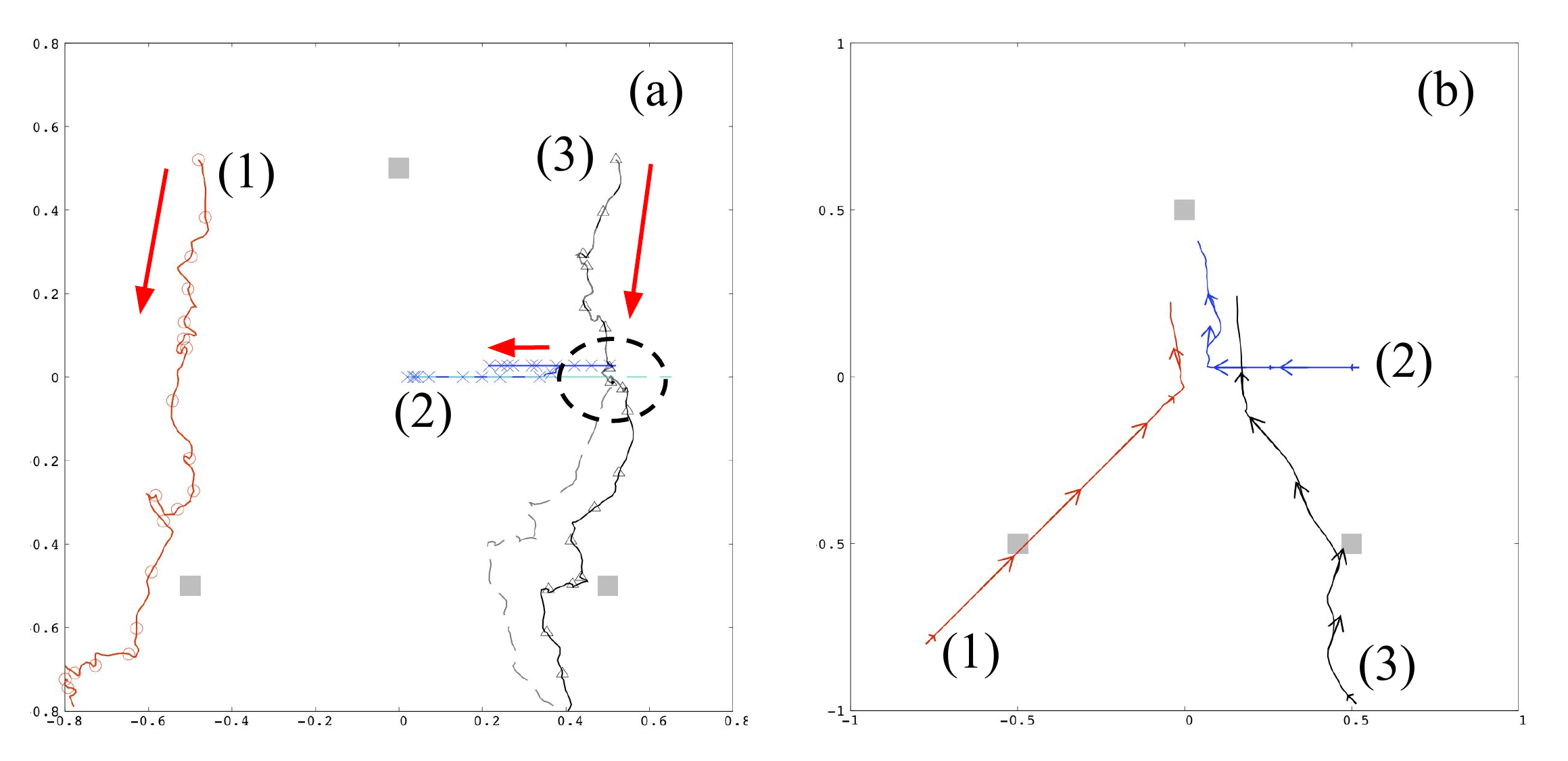}
  \caption{
  Agent trajectories are driven by 
  each operation plan consequently generated via 
  reinforcement learning [with the MLP
    neural network structure], 
  $Q_\alpha$ first [panel (a)] 
  and $Q_\beta$ [panel (b)]. 
  The trajectories in (a) are the 
  virtual states, ${\bm R}^{\left(h_{s,a}\right)}$ (two-fold), 
  branching for Agent \#2 with respect to hypothesis. 
  Those given in (b) are the real trajectories 
  as obtained via Eq.~(\ref{realized02}).
  The labels (1)-(3) indicate the agents, 
which move along the directions denoted by 
red arrows. 
Dotted circles indicate collisions 
between agents. 
  }
  \label{realTraj}
\end{figure*}

\vspace{2mm}
The learning process $Q_\alpha$ in step [1] is performed 
using two virtual spaces
$V^{\left(h_s,a\right)}$, 
corresponding to the two hypotheses 
\begin{eqnarray}
{\bm R^{\left(h_l\right)}}\in V^{\left(h_l\right)}
\ . 
\end{eqnarray}
Each ${\bm R^{\left(h_l\right)}}$ can take 
such possibilities under each constraint 
of its hypothesis ({\it e.g.}, 
$y_3$ cannot be updated due to the 
actuator error). 
For an operation ${\bm g}$, 
the state on each virtual space is 
updated as 
\begin{eqnarray}
{\bm g}:
\begin{array}{c}
  {\bm R}^{\left(h_s\right)}
  \to \tilde{\bm R}^{\left(h_s\right)}\left({\bm g},{\bm R}^{\left(h_s\right)}\right) \\ 
      {\bm R}^{\left(h_a\right)}
      \to \tilde{\bm R}^{\left(h_a\right)}\left({\bm g},{\bm R}^{\left(h_a\right)}\right) \\ 
\end{array}
\ .
\label{twoVirtual}
\end{eqnarray}
Taking the value function, 
\begin{align}
&\rho^{\left(\alpha\right)}\left({\bm g},{\bm R^{\left(h_1\right)}},{\bm R^{\left(h_2\right)}}\right)\nonumber\\
&={\left\|
\tilde{\bm R}^{\left(h_1\right)}\left({\bm g},{\bm R^{\left(h_1\right)}}\right)
-
\tilde{\bm R}^{\left(h_{2}\right)}\left({\bm g},{\bm R^{\left(h_1\right)}}\right)
\right\|} 
\ ,
\nonumber \\
\label{evaluationAlpha}
\end{align}
the two-fold $Q$-table is updated self-consistently as 
\begin{eqnarray*}
  &&
  Q\left({\bm g},{\bm R^{\left(h_1\right)}},{\bm R^{\left(h_2\right)}}\right)\\
&&=\rho^{\left(\alpha\right)}\left({\bm g},{\bm R^{\left(h_1\right)}},{\bm R^{\left(h_2\right)}}\right)
\nonumber \\
&&\quad+\sum_{{\bm g'},{\bm R'^{\left(h_1\right)}},{\bm R'^{\left(h_2\right)}}}
F\left(
\left\{Q\left({\bm g'},{\bm R'^{\left(h_1\right)}},
{\bm R'^{\left(h_2\right)}}\right)\right\}, \right. \\
&&\quad\quad\quad\quad\quad\quad\quad\quad\quad\quad\quad \left.
\left\{\pi\left({\bm g'},{\bm R'^{\left(h_1\right)}},
{\bm R'^{\left(h_2\right)}}\right)\right\}\right) 
\ .
\end{eqnarray*}
Denoting the converged table as 
$\bar{Q}_\alpha\left({\bm g},{\bm R^{\left(h_1\right)}},{\bm R^{\left(h_2\right)}}\right)$, 
the sequence of operations is 
obtained as given in Eq.(\ref{sequence}); in other words, 
\begin{eqnarray}
\left\{
\bar{\bm g}_0^{\left(\alpha\right)},
\bar{\bm g}_1^{\left(\alpha\right)},
\cdots
\bar{\bm g}_M^{\left(\alpha\right)}
\right\}
\label{plan01}
\ .
\end{eqnarray}
The operation sequence generates 
the two-fold sequence of 
(virtual) state evolutions as 
\begin{eqnarray}
\left\{
{\bm R}_1^{\left(h_{s,a}\right)}\to
{\bm R}_2^{\left(h_{s,a}\right)}\to
\cdots\to
{\bm R}_M^{\left(h_{s,a}\right)}
\right\}
\ , 
\label{expected01}
\end{eqnarray}
as shown in Fig.~\ref{realTraj}(a). 

\vspace{2mm}
In Step [2], the agents operate 
according to the plan expressed by Eq.(\ref{plan01}) 
to update (real) states as 
\begin{eqnarray}
\left\{
{\bm R}_1\to
{\bm R}_2\to
\cdots\to
{\bm R}_M
\right\}
\ ,
\label{observed01}
\end{eqnarray}
to be observed by the command base. 
The base compares Eqs.~(\ref{observed01}) and 
(\ref{expected01}) to identify 
whether $h_s$ or $h_a$ is the cause of failure 
($h_a$ in this case).

\vspace{2mm}
{
In Step [3], $Q_\beta$-learning is performed for reward 
$\rho^{\left(\beta\right)}$. 
The reward function $\rho^{\left(\beta\right)}$ calculates 
the sum of the individual agents' rewards, 
where each agent gets a reward of 
$a/(r+1) + b\cdot\delta\left(r\right)$ 
depending on its distance $r$ from the goal post. 
Thus, a higher reward is realized when 
the agent gets closer to the goal post. 
By setting $a=0.01$ and $b=100.0$, 
a much higher reward value ($a+b$) is obtained 
when the agent reaches the goal post ($r=0$). 
Although learning efficiency varies depending on 
the values of $a$ and $b$, 
a relatively high efficiency was achieved by setting $b\gg a$. 
}
The operation sequence is then obtained as 
\begin{eqnarray}
\left\{
\bar{\bm g}_{M+1}^{\left(\beta\right)},
\bar{\bm g}_{M+2}^{\left(\beta\right)},
\cdots
\bar{\bm g}_{L}^{\left(\beta\right)}
\right\}
\ ,
\label{plan02}
\end{eqnarray}
by which the states of the agents are 
updated as 
\begin{eqnarray}
\left\{
{\bm R}_{M+1}\to
{\bm R}_{M+2}\to
\cdots\to
{\bm R}_L
\right\}
\ , 
\label{realized02}
\end{eqnarray}
as shown in Fig.~\ref{realTraj}(b). 

\section{Discussions}
\label{sec.discussions}
Fig.~\ref{realTraj}(a) depicts two-fold 
trajectories, Eq.~(\ref{expected01}), 
corresponding to 
the hypotheses $h_a$ and $h_s$. 
While 
${\bm R}^{\left(h_a\right)}
={\bm R}^{\left(h_s\right)}$ for Agent \#1, 
the branching 
${\bm R}^{\left(h_a\right)}
\ne {\bm R}^{\left(h_s\right)}$ occurs 
for Agent \#2 during operations. 
The branching process earns a score 
via the value function 
$\rho^{\left(\alpha\right)}$ in Eq.~(\ref{evaluationAlpha}), 
which indicates that the learning $Q_\alpha$ was 
conducted properly. Thus, the ability to capture the difference between 
$h_a$ and $h_s$ has been realized.
The red dotted circle shown in (a) 
represents a collision between Agents \#2 and \#3, 
inducing the difference between 
${\bm R}^{\left(h_a\right)}$ and 
${\bm R}^{\left(h_s\right)}$ 
(the trajectories only reflect 
the central positions of agents, while 
each agent has a finite radius similar to its 
size; therefore, the trajectories themselves 
do not intersect even when a collision 
occurs). 
In addition, the collision strategy 
is never generated in a rule-based manner, as 
the agents autonomously deduce their strategy
via reinforcement learning. 

\vspace{2mm}
Three square symbols (closed) situated 
at the edges of a triangle in Fig.~\ref{realTraj} 
represent the goalposts for the conveying mission. 
Fig.~\ref{realTraj}(b) shows the real trajectories 
for the mission, where the initial locations of 
the agents are the final locations in the panel (a). 
From their initial locations, 
Agents \#1 and \#3 immediately arrived 
at their goals to complete each mission, 
and subsequently headed to 
Agent \#2 for assistance.  
Meanwhile, Agent \#2 attempted to reach its goal 
using its limited mobility; 
{\it i.e.,} only along the $x$-axis. 
At the closest position, all three agents 
Coalesced, and Agents \#1 and \#3 began pushing Agent \#2 
up toward the goal. 
Though this behavior is simply the consequence 
of earning more from the value function 
$\rho^{\left(\beta\right)}$, 
it appears as if Agent \#1 wants to 
assist the malfunctioning agent cooperatively 
{
(a video of the behavior shown in 
Fig.~\ref{realTraj}(b) is available 
at the link~\cite{movie}).?
}
By identifying the constraint $h_a$ for 
the agents in the learning phase $Q_\alpha$, 
the subsequent learning phase $Q_\beta$ 
is confirmed to generate the optimal 
operation plans to ensure that the team 
maximizes their benefit through
cooperative behavior as if an autonomous 
decision has been made by the team. 

\begin{figure*}[htbp]
  \includegraphics[width=\hsize]{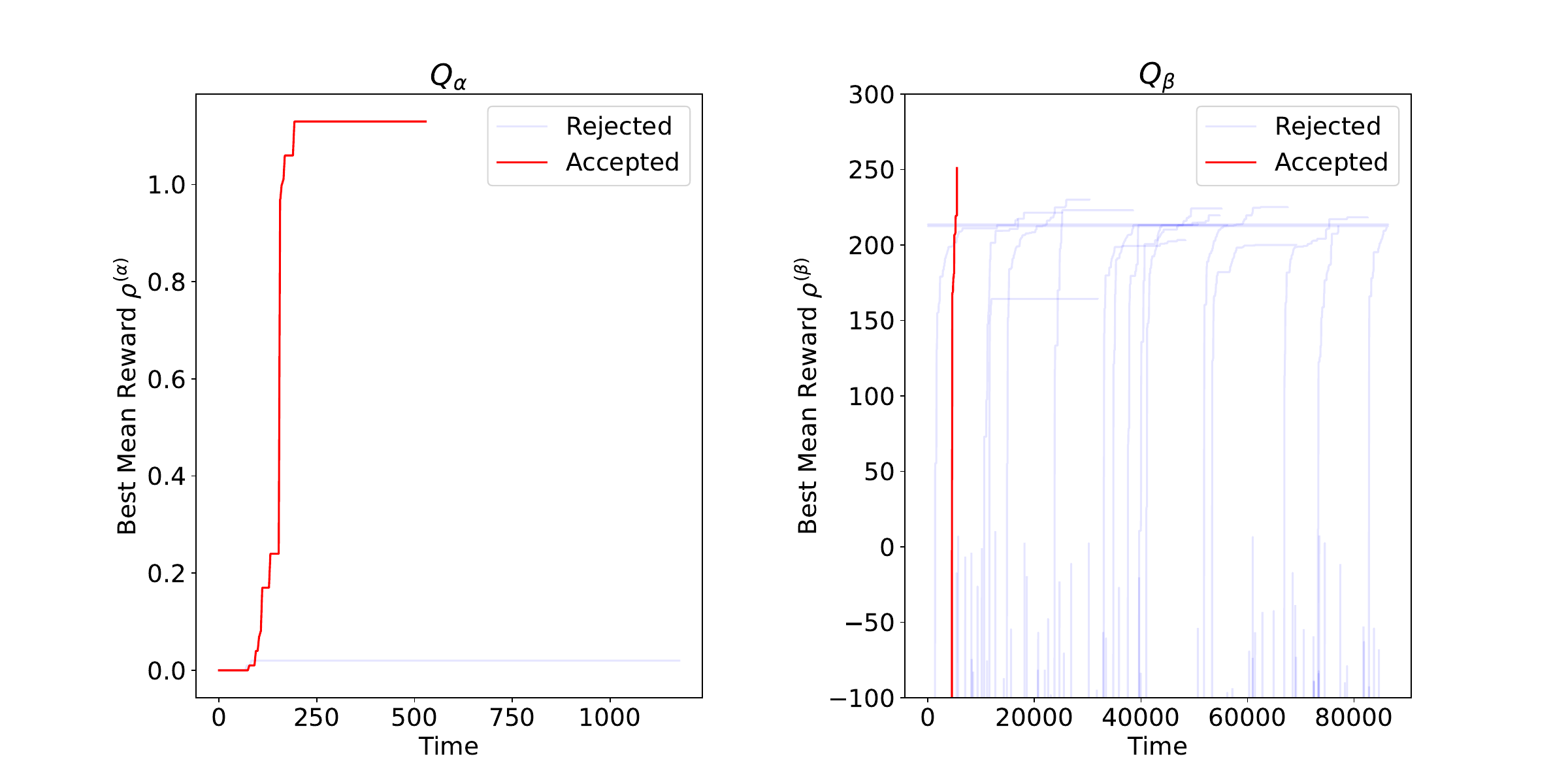}
  \caption{
{  
Learning curve evaluated for MLP network construction 
in terms of the reward function.  
Results for $Q_\alpha$ (left panel) and 
$Q_\beta$ (right panel) are shown. 
Blue and red curves correspond to 
trajectories that did not reach the target reward 
at the end of training, and those that successfully 
reached the target reward, respectively.  
}
  }
  \label{mlptraj}
\end{figure*}

\vspace{2mm}
{
During training, 
if the target reward is not reached in 
the given number of training sessions, 
the training process is reset to avoid being trapped
by the local solution. 
In Fig.~\ref{mlptraj}, 
the training curves of rejected 
trials are shown in blue, whereas 
the acceptable result is shown in red. 
Evidently, more learning processes were rejected 
in $Q_{\beta}$ (right panel) than in $Q_{\alpha}$ (left panel). 
This indicates that it is a more challenging task 
to perform transport planning with three malfunctioning agents,
than to plan the action to 
pinpoint a hypothesis among two. 
However, under more complex failure conditions, 
more learning is expected to be rejected 
for $Q_{\alpha}$ as well, 
as the number of possible hypotheses increases. 
}

\vspace{2mm}
LSTM and MLP were compared in performance 
in terms of the success rate for obtaining working trajectories to distinguish between the hypotheses. 
Notably, even when applying the well-converged
$Q$-table, there is a certain rate required for the
non-working trajectories to eliminate the
difference between the hypotheses.
This is a result of the stochastic nature
of the policy Eq.~(\ref{policyProb})
in generating the trajectories. 
In the present work, we took 
50 independent $Q$-tables, each of which 
was generated from scratch, and obtained 
50 corresponding trajectories. 
The rate required to obtain the trajectories 
required to distinguish among the hypotheses amounts to 94\% for LMS, 
and 78\% for LSTM. 
In the present comparison, 
we used the same iteration steps 
as for $Q$-table convergence. 
Because LSTM has a more complex internal 
structure, its learning quality was
expected to be relatively lower than that of LMS 
for the common condition, and its performance rate was likewise expected to be lower. 
In other words, a higher iteration cost 
is required for LSTM to achieve 
performance comparable to LMS. 
{
As such, the results shown in the main text 
are those obtained by LMS, whereas 
those obtained by LSTM are presented in the Appendix for 
reference.
}

\vspace{2mm}
For a simulation in a virtual 
environment space, 
we must evaluate the distances 
between agents at every step. 
As this is a pairwise evaluation, 
its computational cost scales as 
$\sim N^2$ for $N$ agents. 
This cost scaling can be mitigated 
by using the domain decomposition method
wherein each agent is evaluated according to its voxel, 
and the distance between agents is represented 
by that between corresponding voxels registered in advance. 
The corresponding cost scales linearly with $N$ 
at a much faster rate than the naive 
$\sim N^2$ evaluation method as the number of agents $N$ 
increases.

\section{Conclusion}
\label{sec.conc}
{
Agents performing group missions can
suffer from errors 
during a missions. 
Multiple hypotheses 
may be devised to explain the causes of such errors. 
Cooperative behaviors, such as 
collisions between agents, can be deployed to identify 
said causes. 
We considered the autonomous planning 
of group behaviors via 
machine-learning techniques. 
Different hypothesizes explaining 
the causes of the errors 
lead to different expected states 
as updated from the same initial state by 
the same operation. 
The larger the difference gets, 
the better the corresponding operation plan is able
to distinguish between the different hypotheses. 
In other words, the magnitude of the difference 
can be the value function to 
optimize the desired operation plan. 
Gradient-based optimization does not 
work well because a tiny fraction 
among the vast possible operations 
({\it e.g.}, collisions) can capture 
the difference, leading to a
sparse distribution of the finite 
value for the function. 
We discovered that 
reinforcement learning is the obvious 
choice to be applied for such problems. 
Notably, the optimal plan 
obtained via reinforcement 
learning was the operation 
that causes agents to collide with each other. 
To identify the causes of error using this plan, 
we developed a revised mission plan that incorporates 
the failure by another learning where 
the malfunctioning agent receives assistance from other agents. 
By identifying the cause of failure, 
the reinforcement learning process plans 
a revised mission plan that considers 
said failure to ensure an appropriate cooperation procedure. 
}

\section{Acknowledgments}
The computations in this work were performed 
using the facilities at 
Research Center for Advanced Computing 
Infrastructure at JAIST. 
R.M. is grateful for financial support from 
MEXT-KAKENHI (19H04692 and 16KK0097), 
from the Air Force Office of Scientific Research 
(AFOSR-AOARD/FA2386-17-1-4049;FA2386-19-1-4015). 
We would like to thank Kosuke Nakano for his feedback,
as it significantly helped improve the overall paper. 

\section{Appendix}
\label{appendix}
\subsection{Policy function}
\label{policy}
Getting the benefit $Q\left(\bm R, \bm g\right)$,
the most naive choice for the next action would be 
\begin{align}
  {\bm g}^*\left(\bm R\right)=\arg\max_{\bm g}Q\left({\bm R},{\bm g}\right)
  \ ,
\label{opt01}
\end{align}
known as the greedy method.
This is represented in terms of the
policy distribution function as 
\begin{align*}
\pi\left(\bm R,\bm g\right)
=\delta\left({\bm g}-{\bm g^*}\left(\bm R\right)\right)
\ .
\end{align*}
To express explicitly that  
${\bm g}^*\left(\bm R\right)$
depends on $Q\left(\bm R,\bm g\right)$
via Eq.~(\ref{opt01}),
let  
\begin{align*}
\pi\left(\bm R,\bm g\right)
=\delta\left({\bm g}-{\bm g^*\left(Q\left(\bm R,\bm g\right)\right)}\right)
=\Delta\left(Q\left(\bm R,\bm g\right)\right)
\ ,
\end{align*}
be a special case of 
\begin{align*}
\pi\left(\bm R,\bm g\right)
=P\left(Q\left(\bm R,\bm g\right)\right)
\ .
\end{align*}

\vspace{2mm}
To improve the probability of obtaining 
the optimal solution compared to the greedy method 
in a delta-function wise distribution, 
there are several choices allowing for 
a finite probability for 
$\bm g \ne \bm g^*$, including the
``$\varepsilon$-greedy method``: 
\begin{align*}
P\left(Q\left(\bm R,\bm g\right)\right)
=\left(1-\varepsilon\right)\cdot\Delta\left(Q\left(\bm R,\bm g\right)\right)
+\varepsilon\cdot\left[{\rm RandomNubmer}\right]
\ ,
\end{align*}
and the ``Boltzmann policy``: 
\begin{align*}
P\left(Q\left(\bm R,\bm g\right)\right)
=\exp\left(-\beta Q\left(\bm R,\bm g\right) \right)
\ .
\end{align*}

\vspace{5mm}
\subsection{Results using LSTM}
\label{res.lstm}
\begin{figure*}[htbp]
  \includegraphics[width=\hsize]{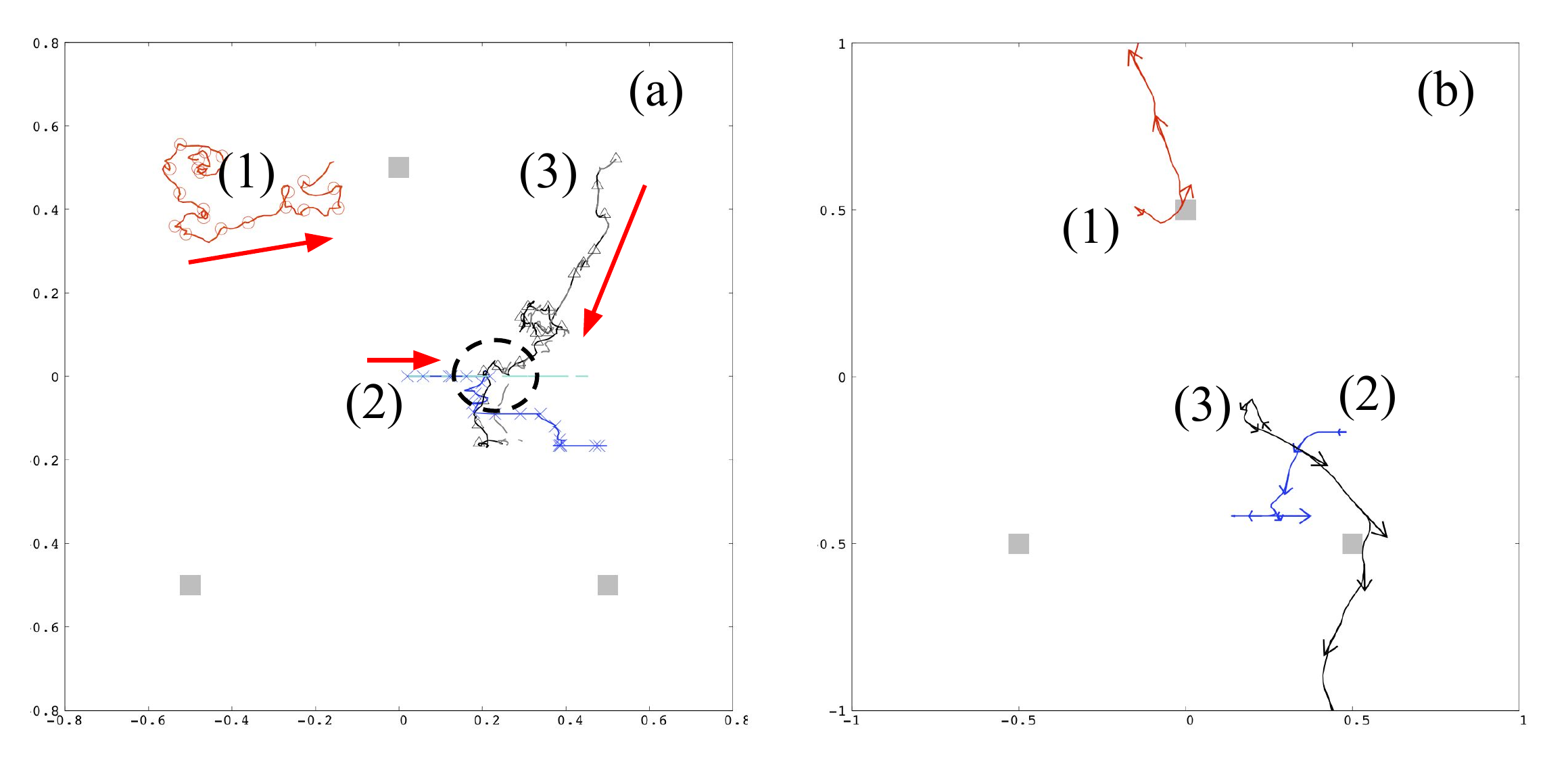}
  \caption{
Agent trajectories driven by 
each operation plan generated via 
reinforcement learning [with LSTM], 
$Q_\alpha$ first [panel (a)] 
and $Q_\beta$ [panel (b)], consequently. 
The trajectories in (a) are the 
virtual states, ${\bm R}^{\left(h_{s,a}\right)}$ (two-fold), 
branching for Agent \#2 with respect to each hypothesis. 
Those given in (b) are the real trajectories 
as given in Eq.~(\ref{realized02}).
The labels (1)-(3) represent agents,
which move along the directions denoted by
red arrows. 
Dotted circles indicate collisions 
between agents.
  }
  \label{realTrajLstm}
\end{figure*}

\begin{figure*}[htbp]
  \includegraphics[width=\hsize]{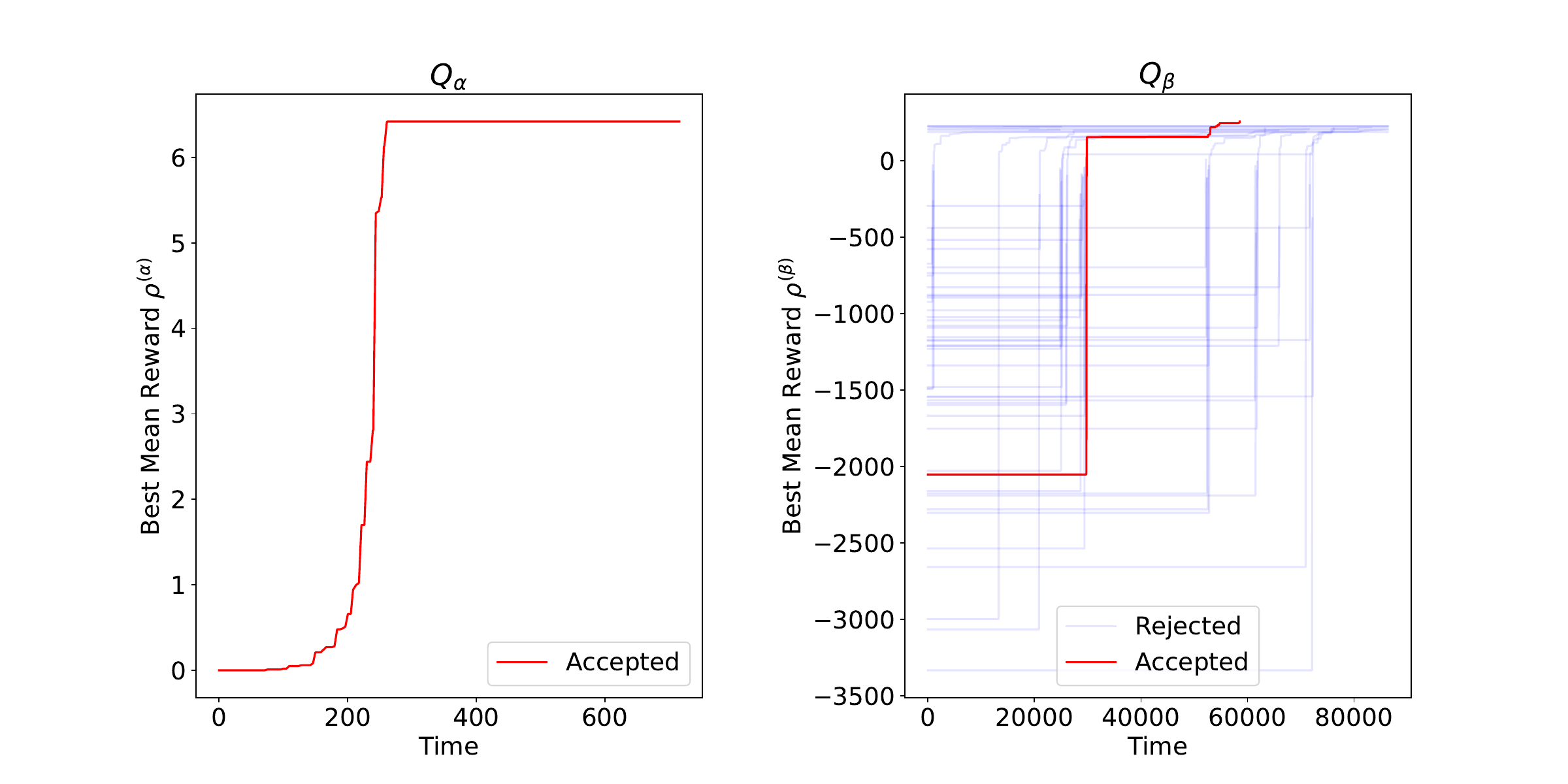}
  \caption{
{  
Learning curve in terms of the reward function, 
evaluated for LSTM network construction.  
Results for $Q_\alpha$ (left panel) and 
$Q_\beta$ (right panel) are shown. 
Blue and red curves correspond to 
trajectories that did not reach the target reward 
at the end of training, and those that successfully 
reached the target reward, respectively.  
}
  }
  \label{lstmtraj}
\end{figure*}

As explained in the main text, 
reinforcement learning using the
LSTM neural network structure 
leads to nearly identical 
behaviors for the agents, 
though it requires a lower rate to obtain 
working trajectories to distinguish 
among the hypotheses. 
{
Figs.~\ref{realTrajLstm} and ~\ref{lstmtraj} 
show the optimized trajectories and learning 
curves, respectively (counterparts 
of Figs.~\ref{realTraj} and ~\ref{mlptraj}, 
respectively). 
}

\bibliography{references}

\end{document}